\documentclass[sigconf]{acmart}
\usepackage{algorithm}
\usepackage{enumitem}

\usepackage{algpseudocode}
\usepackage{amsmath}
\usepackage{amsthm}
\usepackage{colortbl}

\newtheoremstyle{mydefinition}
  {6pt}                    
  {6pt}                    
  {\normalfont}            
  {}                       
  {\bfseries}              
  {.}                      
  {0.5em}                  
  {\thmname{#1}\thmnumber{ #2}\thmnote{ (#3)}}

\theoremstyle{mydefinition}
\newtheorem{definition}{Definition}
\usepackage{amsfonts}
\usepackage{amssymb}
\usepackage{booktabs}
\usepackage{multirow}
\usepackage{graphicx}

\usepackage{caption}
\AtBeginDocument{%
  }
\setcopyright{acmlicensed}
\copyrightyear{2026}
\acmYear{2026}
\setcopyright{cc}
\setcctype{by}
\acmConference[SIGSPATIAL '26]{The 34th ACM International Conference on Advances in Geographic Information Systems}{November 03--06, 2026}{Riverside, CA, USA}
\acmBooktitle{The 34th ACM International Conference on Advances in Geographic Information Systems (SIGSPATIAL '26), November 03--06, 2026, Riverside, CA, USA}
\acmDOI{10.1145/3841645.3842986}
\acmISBN{979-8-4007-2950-8/2026/11}
\acmSubmissionID{143}

\begin{document}
\widowpenalty=0
\clubpenalty=0
\title{Reinforcement Learning for UAV-Based Participatory Sensing 
in Dynamic Environments}


\author{Xin Ouyang}
\affiliation{%
  \institution{The Hong Kong University of Science and Technology (Guangzhou)}
  \city{Guangzhou}
  \country{China}}
\email{xouyang755@connect.hkust-gz.edu.cn}

\author{Songxin Lei}
\affiliation{%
  \institution{The Hong Kong University of Science and Technology (Guangzhou)}
  \city{Guangzhou}
  \country{China}}
\email{slei924@connect.hkust-gz.edu.cn}

\author{Xusen Guo}
\affiliation{%
  \institution{The Hong Kong University of Science and Technology (Guangzhou)}
  \city{Guangzhou}
  \country{China}}
\email{xgou796@connect.hkust-gz.edu.cn}

\author{Yutian Jiang}
\affiliation{%
  \institution{The Hong Kong University of Science and Technology (Guangzhou)}
  \city{Guangzhou}
  \country{China}}
\email{yjiang194@connect.hkust-gz.edu.cn}

\author{Sijie Ruan}
\affiliation{%
  \institution{Beijing Institute of Technology}
  \city{Beijing}
  \country{China}}
\email{sjruan@bit.edu.cn}

\author{Yuxuan Liang}
\authornote{Corresponding author.}
\affiliation{%
  \institution{The Hong Kong University of Science and Technology (Guangzhou)}
  \city{Guangzhou}
  \country{China}}
\email{yuxliang@outlook.com}

\renewcommand{\shortauthors}{Ouyang et al.}

\begin{abstract}
Using Unmanned Aerial Vehicle (UAV) for urban sensing has emerged as a powerful paradigm to monitor the status of the city, e.g., air quality and noise levels, through agile aerial crowdsourcing. Despite this potential, existing UAV-based sensing approaches overlook environmental disturbances like wind that drastically impact drone velocity and energy efficiency. Consequently, directly applying existing methods to this joint delivery and sensing paradigm in dynamic environments faces two severe challenges: (1) scalability bottlenecks as fleet sizes expand; and (2) multi-timescale decision heterogeneity between macro task dispatching and micro velocity control. To tackle these, we formalize the problem as SensUAV and propose a \underline{T}wo Time\underline{S}cale \underline{R}einforcement \underline{L}earning framework (\textbf{TSRL}). Specifically, TSRL separates decision-making into two cooperative layers. At the macro level, a task-embedding sensing dispatcher handles scalability by separately encoding distinct task features and sequentially evaluating UAV suitability before task selection. At the micro level, a wind-aware velocity controller learns fine-grained velocity scheduling to adapt to dynamic environmental variations. Extensive experiments on real-world datasets demonstrate that TSRL significantly outperforms baselines, achieving average system profit improvements of 20.1\% in Hangzhou and 46.6\% in Shanghai. The source code is available at \url{https://github.com/awhiteawhite/TSRL_SensUAV}.
\end{abstract}

\ccsdesc[500]{Information systems~Spatial-temporal systems}
\ccsdesc[300]{Applied computing~Multi-criterion optimization and decision-making}
\keywords{Urban Sensing, Spatial Crowdsourcing, Reinforcement Learning, Urban Computing, Spatio-temporal Data Mining}



\maketitle
\vspace{-2mm}
\section{Introduction}
Urban sensing plays a fundamental role in smart-city systems by continuously collecting spatial and temporal information for traffic monitoring, environmental surveillance, emergency response, and public service optimization~\cite{chen2024ddl,zheng2019urban}. Traditional ground-based participatory sensing relies on mobile users, vehicles, or fixed infrastructures to collect urban data ~\cite{guo2026agentsense,ji2016urban,wang2024urban}. Although effective in many scenarios, ground-based sensing is inherently constrained by road networks, uneven participant distribution, limited spatial coverage, and delayed response to emerging sensing demands.

\begin{figure}[t]
    \centering
    \includegraphics[width=0.95\columnwidth]{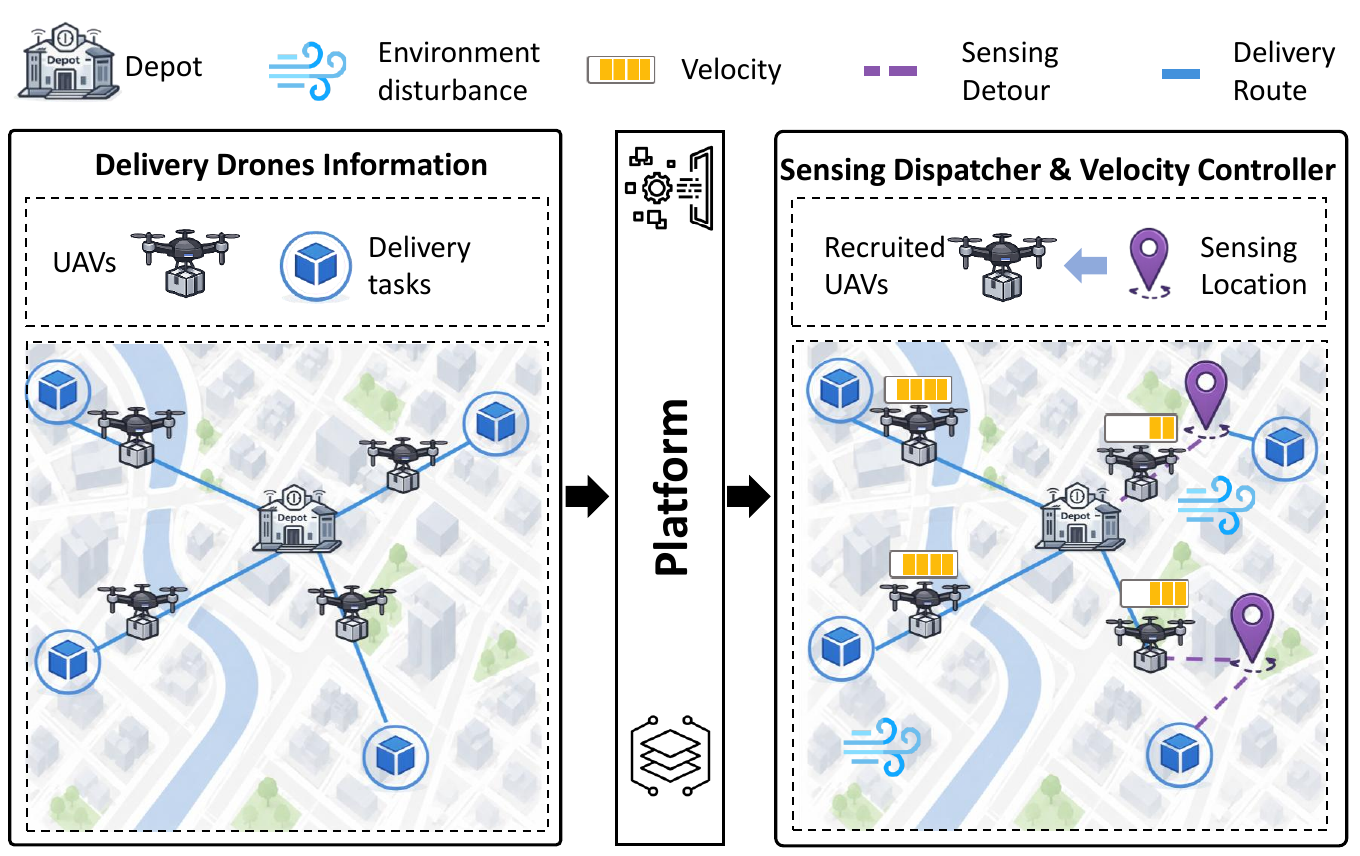} 
    \caption{Participatory Sensing for Delivery Drones}
    \label{fig:intro}
\end{figure}

Recent advances in unmanned aerial vehicles (UAVs) provide a promising alternative for large-scale urban sensing. Compared with ground-based platforms, UAVs can actively move toward sensing hotspots and support flexible deployment in complex urban environments ~\cite{wang2023air}. Meanwhile, the rapid development of low-altitude logistics is creating large fleets of delivery drones that routinely travel across urban areas. As sensing participants, delivery drones are less constrained by road topology than ground participants, and they can piggyback sensing tasks on existing logistics operations instead of requiring dedicated sensing-only UAV flights, thereby reducing the additional operational cost of urban data collection. Therefore, participatory sensing for delivery drones is expected to become a key component of future urban data acquisition systems~\cite{liu2025delay,chen2024ddl}.
\vspace{-0.9mm}
\\
\indent Early research on UAV-based urban sensing has made steady progress in heuristic-based static optimization and reinforcement learning approaches~\cite{xiang2021reusing,liu2025delay,chen2024ddl}, which typically focus on solving combinatorial allocation problems or dynamically optimizing \mbox{decisions} regarding sensing assignment. However, many works assume fixed flight speed and distance-based travel cost~\cite{chen2022deliversense,xiang2021reusing,lei2026hierarchical}, which overlook the fact that UAV velocity is strongly affected by environmental disturbances (wind conditions) and the distance to the target point. In practice, wind changes the effective flight speed and power consumption of a UAV, causing the same sensing detour to incur different travel times and energy costs under different weather conditions. In this paper, we therefore study the problem of Participatory Sensing for Delivery Drones (\textbf{SensUAV}) in dynamic urban environments, where wind conditions continuously affect UAV flight speed and energy consumption. This allows existing delivery drones to be utilized as mobile sensing carriers without being separated from their primary delivery service. More specifically, as illustrated in Figure~\ref{fig:intro}, delivery drones first receive delivery tasks from the depot, and the platform coordinates suitable UAVs to complete sensing tasks during delivery. The platform then jointly determines sensing detours and UAV velocities  with the goal of maximizing overall system profit.

The incorporation of sensing dispatching and velocity scheduling transforms SensUAV into a challenging online mixed-integer sequential optimization task~\cite{bertsimas2022online}. Conventional exact optimization methods can provide high-quality or even optimal solutions for a given deterministic instance, but they are difficult to deploy directly in SensUAV. The system state is continuously updated by online order arrivals, expiring sensing tasks, changing UAV availability, and stochastic weather conditions; consequently, an exact solver would need to repeatedly solve a large mixed-integer sequential optimization problem under strict real-time constraints~\cite{wang2023learning,wang2024learning,wesselmann2012implementing}. Meanwhile, Heuristic and meta-heuristic methods improve online efficiency, but they often rely on hand-crafted rules or short-horizon search, making them less effective in capturing the long-term effects of dispatch and velocity decisions~\cite{chen2025sustainability}. Interacting with the dynamic environment, reinforcement learning provides a practical alternative by learning an offline policy that optimizes long-term discounted system profit and produces fast online decisions via a single forward pass.

However, directly applying existing RL methods faces two challenges to tackle this SensUAV problem:


\textbf{Scalability bottleneck} as the UAV fleet size expands. In large-scale crowdsensing, the rapid expansion of UAVs and heterogeneous tasks introduces severe computational bottlenecks. Conventional Multi-Agent RL paradigms struggle with joint state-space explosion and heavy communication overhead required to avoid task collisions. Alternatively, models combining GNNs with Pointer Networks rely on autoregressive decoding, which forces strictly sequential execution rather than parallel processing\cite{wang2024urban,wang2023learning}.
Consequently, both approaches often incur significant inference latency that severely hinders their real-time deployment.

\textbf{Multi-timescale decision heterogeneity} between dispatch and velocity control.
Sensing dispatch is a relatively persistent high-level decision, whereas velocity control is a fast execution-level decision that should adapt frequently to wind, battery status, deadlines, and target distance. Treating these two decisions at the same temporal resolution either makes dispatch overly reactive or makes velocity control too coarse, while generic hybrid-action RL couples discrete dispatch and continuous speed control in a single mixed action space, leading to noisy credit assignment~\cite{dou2025scheduling}.
Although hierarchical RL provides a natural abstraction, jointly learning the dispatcher and the controller makes the macro-level transition dynamics non-stationary, because the low-level execution policy keeps changing during upper-level training~\cite{hutsebaut2022hierarchical}.    

To tackle these challenges, we proposed a framework called \underline{T}wo Time\underline{S}cale \underline{R}einforcement \underline{L}earning framework (\textbf{TSRL}). To address the first challenge, TSRL employs specific sensing dispatcher. Instead of directly scoring all UAV sensing pairs as unrelated actions, the dispatcher first evaluates UAV-level suitability and then selects a sensing task conditioned on the chosen UAV. This allows the policy to evaluate sensing decisions under the current delivery-service context rather than treating sensing tasks as standalone assignments. To address the second challenge, TSRL separates dispatch and execution into two temporal layers. Velocity controller learns wind-aware velocity scheduling over micro steps, while the upper-level dispatcher acts only at macro decision epochs. 

Our main contributions are summarized as follows:
\begin{itemize}[leftmargin=*,topsep=1pt,itemsep=0pt]
    \item To the best of our knowledge, we are the first to introduce a dynamic-environment-aware participatory sensing system for delivery drones, featuring explicit modeling of wind-aware energy and velocity control. 
    \item We introduce a two-timescale RL framework named TSRL that separates slow sensing assignment from fast motion control. To handle the scalability bottleneck, TSRL designs a specific task embedding sensing dispatcher that first evaluates UAV suitability and then selects a sensing task conditioned on the chosen UAV. To handle heterogeneous timescales, TSRL trains a wind-aware velocity controller at the micro scale. 
    \item Extensive experiments on real-world Hangzhou and Shanghai datasets show that TSRL consistently outperforms representative baselines in system profit. TSRL achieves average profit improvements of $20.1\%$ in Hangzhou and $46.6\%$ in Shanghai over the strongest baseline, with the maximum gains reaching $25.2\%$ and $51.2\%$, respectively.
\end{itemize}

\section{Preliminary}
\subsection{Definitions}

\begin{definition}[Delivery Task]
A delivery task refers to transporting goods from an origin location to a destination location. Specifically, the origin for all delivery tasks is fixed at the central UAV depot within the area. Let the set of delivery tasks be denoted as $D = \{d_1, d_2, \dots, d_{n}\}$, where each delivery task $d_c$ is represented as a tuple $d_c = \langle \mathrm{lng}_c, \mathrm{lat}_c, e_c, l_c \rangle$. $( \mathrm{lng}_c, \mathrm{lat}_c )$ specifies the geographical coordinates of the order's destination. $[e_c, l_c]$ represent the earliest and latest acceptable delivery times for the task, respectively.
\end{definition}

\begin{definition}[UAV Team]
The UAV delivery team is denoted as $ U = \{u_1, u_2, \dots, u_m\}$,
where each UAV is represented as $u_i = \langle \mathrm{lng}_i, \mathrm{lat}_i ,v_{i}, e_{i} \rangle$ .
Here, $\mathrm{lng}_i, \mathrm{lat}_i$ is the current location of UAV $i$, $v_{i}$ and $e_{i}$ denote the flight velocity and remaining energy of UAV $i$, respectively.
\end{definition}
\begin{definition}[Sensing Task]
An urban sensing task, such as environmental monitoring or aerial photography, is represented as $ Z = \{z_1, z_2, \dots, z_k\}$.
Each sensing task $z_j = \langle \mathrm{lng}_j, \mathrm{lat}_j, \mathrm{tw}_s, \mathrm{tw}_e \rangle$,
where $(\mathrm{lng}_j, \mathrm{lat}_j)$ denotes the geographic coordinates of task $j$, and $[\mathrm{tw}_s, \mathrm{tw}_e]$ represents its valid time window.
\end{definition}

\begin{definition}[Hierarchical Entropy-based Data Coverage]
Inspired by prior urban sensing studies ~\cite{ji2016urban}, the sensing coverage metric is defined as a weighted combination of spatial entropy and data quality:
\begin{equation}
\phi(Z) = \alpha E(Z) + (1 - \alpha)\log_2 Q(Z),
\end{equation}
where $Z$ denotes the completed sensing task set, $\phi(Z)$ represents the hierarchical coverage score, $\alpha \in [0,1]$ is the balancing parameter, $E(Z)$ denotes the spatial entropy, and $Q(Z)$ is the number of completed sensing tasks. In this work, we set $\alpha = 0.5$, following \cite{guo2026agentsense}. This entropy-based formulation is adopted to encourage balanced spatial-temporal urban sensing coverage.
\end{definition}

\begin{definition}[Energy Consumption]
In dynamic urban environments, UAV energy consumption is significantly affected by meteorological conditions such as wind. Inspired by prior studies on UAV energy modeling and control ~\cite{gao2020energy,zheng2022joint,ma2024deep}, we model the UAV power consumption using two components: blade profile energy and parasite energy.

Let $\mathbf{w}_t$ denote the wind velocity at time $t$, and let $\mathbf{v}_i$ denote the flight velocity of UAV $i$. The power consumption per minute of UAV $i$ is modeled as
\begin{equation}
P_i = K_1 \|\mathbf{v}_i\|^2 + K_2 \|\mathbf{v}_i - \mathbf{w}_t\|^3,
\end{equation}
where the term $K_1 \|\mathbf{v}_i\|^2$ represents the blade profile energy, while $K_2 \|\mathbf{v}_i - \mathbf{w}_t\|^3$ corresponds to the parasite energy caused by aerodynamic drag and wind resistance. The coefficients $K_1$ and $K_2$ are set as $10^{-5}\,\mathrm{J}\cdot\mathrm{min}^2/\mathrm{m}^2$ and $10^{-9}\,\mathrm{J}\cdot\mathrm{min}^3/\mathrm{m}^3$, respectively, following ~\cite{gao2020energy,zheng2022joint}.

Accordingly, the total energy consumption of UAV $i$ during time interval $\Delta t$ is defined as $ E_i = P_i \Delta t$.
To satisfy safety return requirements, a UAV is forced to return to the charging station once its remaining energy falls below 20\% of the total battery capacity.
\end{definition}


 \subsection{Problem Statement}
\noindent
\textbf{Objective function.} The SensUAV problem aims to jointly optimize sensing dispatch and velocity scheduling to maximize the overall system profit. Overall system profit is formulated as :
\begin{equation}
    J = w_z\phi(Z) + w_o O - w_e E
    \label{eq:overall_profit}
\end{equation}
where $\phi(Z) $ is the sensing entropy, $O$ is the number of completed delivery orders, and $E$ is the total UAV energy consumption. The coefficients $w_z$, $w_o$, and $w_e$ are constant profit weights. In this work, we set $w_z$ to 2000 cents per entropy advancement, which is approximately equal to 500 cents per sensing task, $w_o$ to 200 cents per completed order, and $w_e$ to $10^{-5}$ cents per Joule, according to industry prices and expert knowledge~\cite{almazroi2023multi, chen2025sustainability,guo2018task}. 

\noindent
\textbf{Problem formulation.}
 We define two types of decision variables. The binary variable $\omega_{i,j,t}\in\{0,1\}$ indicates whether sensing task $j$ is inserted into the task list of UAV $i$ at time $t$, and the continuous variable $v_{i,t}$ denotes the flight velocity of UAV $i$ at time $t$. :
\begin{equation}
\begin{aligned}
    \max_{\{\omega_{i,j,t}, v_{i,t}\}} \quad
    & J = w_z\phi(Z) + w_oO - w_eE \\
    \mathrm{s.t.} \quad
    C_1:\;& \omega_{i,j,t} \in \{0,1\},
    && \forall i\in U, j\in Z, t, \\
    C_2:\;& 0 \leq \|v_{i,t}\| \leq v_i^{\max},
    && \forall i\in U, t, \\
    C_3:\;& e_{i,t+1} = e_{i,t} - P_i(v_{i,t}, w_t)\Delta t,
    && \forall i\in U, t, \\
    C_4:\;& e_{i,t} \geq e_i^{\mathrm{return}}(p_{i,t}),
    && \forall i\in U, t, \\
    C_5:\;& \sum_{i\in U}\sum_t \omega_{i,j,t} \leq 1,
    && \forall j\in Z .
\end{aligned}
\label{eq:sensuav_formulation}
\end{equation}
Here, $v_i^{\max}$ is the maximum velocity of UAV $i$, $e_{i,t}$ is its remaining energy at time $t$, and $e_i^{\mathrm{return}}(p_{i,t})$ denotes the minimum energy required for UAV $i$ to safely return from its current position $p_{i,t}$ to the depot. $P_i(v_{i,t}, w_t)$ denotes the wind-aware power consumption of UAV $i$ under velocity $v_{i,t}$ and wind condition $w_t$, and $\Delta t$ is the duration of one micro control step. Constraint $C_1$ defines the binary sensing-insertion decision, $C_2$ enforces the UAV speed limit, $C_3$ describes the wind-aware energy dynamics, and $C_4$ guarantees safe return to the depot. Constraint $C_5$ prevents duplicate sensing by ensuring that each sensing task is assigned at most once.
\\
\noindent
\textbf{Lemma 1. The problem is $NP-hard$}. 

\noindent
\textit{Proof.} We prove this by reducing this problem into a 0-1 knapsack problem, which is already known as a NP-hard problem. 
We first give the definition of 0-1 knapsack problem: 
Assume that there is a set of items with a weight and a value identically, and a knapsack. We need to determine which item should be put into the knapsack, so that the total weight can satisfy the constraint and the total value can be as large as possible.

Here we can provide a special case of our SensUAV problem. Consider only a single UAV and a fixed delivery route{depot - destination - depot}, and a fixed flight speed. The only remaining decision is the sensing task should be inserted to which predetermined route. Here for this specific subset problem, we create a sensing task whose insertion causes an additional cost of energy. To solve this problem, we want to find an insertion policy to satisfy the energy constraint meanwhile maximizing the sensing entropy. Therefore, selecting a feasible subset of sensing tasks with objective gain at least is a 0-1 knapsack problem. Therefore, this restricted problem is NP-hard, and the general SensUAV problem is also NP-hard.



\begin{figure*}[t]  
    \centering
    \includegraphics[width=1.00\textwidth]{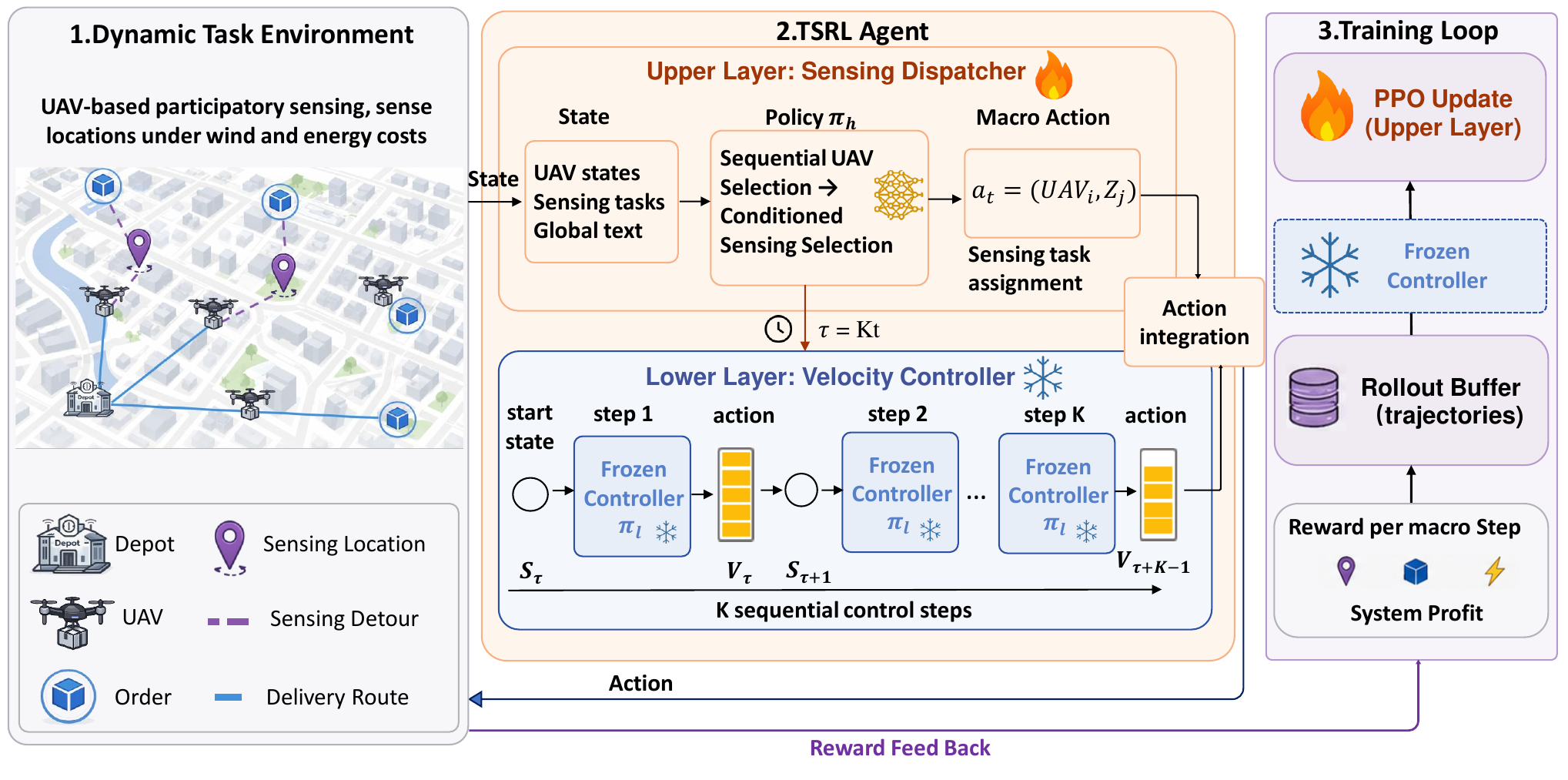}  
    \caption{Framework for SensUAV}
    \label{fig:uav_framework}
\vspace{-2mm}
\end{figure*}

\section{Method}

\subsection{Overview}
we propose a two-stage reinforcement learning framework, termed \textbf{TSRL}, to maximize the system profit through joint sensing-task dispatch and UAV motion control. As illustrated in Fig.~\ref{fig:uav_framework}, TSRL adopts a two-timescale framework to decouple two heterogeneous decisions, namely sensing-task dispatch and UAV velocity control. Specifically, the upper-level \emph{Sensing Task Dispatcher} operates at the macro timescale and determines whether sensing tasks should be inserted into ongoing delivery routes, while the lower-level \emph{Velocity Controller} operates at the micro timescale and continuously outputs UAV velocity actions under dynamic environmental conditions.

During system execution, delivery orders dynamically arrive at depot stations and are assigned using a first-in-first-out (FIFO) policy. At the end of each micro control step, the system checks whether a UAV has returned to a depot station. If the depot contains pending delivery orders and the UAV has sufficient remaining battery, the earliest pending order is assigned to the UAV for execution. Meanwhile, UAVs may opportunistically conduct sensing tasks during the delivery process. One macro decision interval contains $K$ micro control steps, enabling long-term sensing dispatch planning while preserving fine-grained motion adaptation. To improve training stability, we first train the low-level velocity controller, after which the high-level sensing dispatcher is optimized on top of the induced two-time-scale environment dynamics. The complete two-stage training procedure is summarized in \ref{alg:two_level_training}.

\vspace{-5mm}
\subsection{Sensing Task Dispatcher}

\subsubsection{\textbf{MDP Formulation}}

We formulate the upper-layer sensing task dispatcher as a macro-level MDP:
\begin{equation}
\mathcal{M}^{h} = (\mathcal{S}^{h}, \mathcal{A}^{h}, \mathcal{P}^{h}, \mathcal{R}^{h}, \gamma),
\end{equation}
where $\mathcal{S}^{h}$ is the macro state space, $\mathcal{A}^{h}$ is the macro action space, $\mathcal{P}^{h}$ is the macro transition function, $\mathcal{R}^{h}$ is the macro reward function, and $\gamma \in (0,1)$ is the discount factor.

Different from a standard single-timescale MDP, the environment evolves at the micro level, while the sensing task dispatcher acts once every $K$ micro steps. Since $K$ is fixed in our system, the upper-layer process remains a discrete-time MDP with fixed transition intervals, for $K$'s value, we will discuss in the hyperparameter experiment part.

\noindent
\textbf{State Space}. At macro step $t$, the dispatcher observes the global system state
\begin{equation}
S_t^{h} = \big(U_t, Z_t, W_t, g_t\big),
\end{equation}
where $U_t$ denotes the current UAV fleet state, $Z_t$ denotes the set of currently available sensing tasks, and $W_t$ denotes the weather condition, including wind speed and wind direction. The optional global context vector $g_t$ summarizes system-level statistics such as current workload, remaining active sensing tasks, and fleet energy status.


More concretely, the macro state augments the basic UAV attributes with runtime execution information. For each UAV, we use its current location, velocity, remaining energy, and task status, where the task status indicates whether the UAV is idle, assigned, delivering, returning, charging, or returning due to low battery. Each sensing task contains its spatial and temporal attributes, including task location and valid time window.

\noindent
\textbf{Action Space}. At each macro step, sensing task dispatcher selects a UAV--sensing pair$ \ a_t^{h} = (u_i, z_j)$, where $\ u_i \in U_t,\ \ z_j \in Z_t \cup \{\varnothing\}$. $\varnothing$ denotes a null action, meaning that no new sensing task is assigned at the current macro step. Thus, dispatcher either assigns sensing task $z_j$ to UAV $u_i$, or skips assignment when no suitable sensing action is beneficial. Equivalently, the action can be represented in index form as
$a_t^{h} = (i,j)$, where $ \ i \in \{1,\dots,M\}, \ j \in \{0,1,\dots,|Z_t|\}$. $j=0$ denotes the null action. Invalid UAVs and inactive sensing tasks are masked out during action sampling.

\noindent
\textbf{Transition Function}. After the dispatcher outputs macro action $a_t^{H}$, the environment first updates the task manager by committing the selected sensing assignment. Then, the lower-layer velocity controller executes $K$ consecutive micro control steps:
\begin{equation}
a_{\tau}^{l} = v_{\tau} = \pi^{l}(s_{\tau}^{l}), \qquad \tau=tK,\dots,(t+1)K-1,
\end{equation}
where $\pi^{l}$ denotes the low-level velocity controller and $s_{\tau}^{l}$ is the micro-level observation.

During these $K$ micro steps, the simulator updates UAV positions, battery levels, delivery progress, sensing completion, and weather-dependent flight dynamics. After the rollout finishes, the environment returns the next macro state $s_{t+1}^{h}$. Therefore, the macro transition is defined as$ \  s_{t+1}^{h} \sim \mathcal{P}^{h}(\cdot \mid s_t^{h}, a_t^{h}) $, $\mathcal{P}^{h}$ is induced by $K$ micro transitions under the velocity controller.

\noindent
\textbf{Reward Function}. The macro reward is designed to reflect the overall system objective while alleviating sparse-reward issues. Specifically, the reward at macro step $t$ is defined as
\begin{equation}
r_t^{h}
=\Delta J
+
r_t^{\mathrm{timeout}}
+
r_t^{\mathrm{shape}},
\label{eq:macro_reward_final}
\end{equation}
where $J = w_z\phi(Z) + w_oO-w_eE $ denotes the system profit, $r_t^{\mathrm{timeout}}$ is a timeout penalty introduced to encourage timely delivery completion and improve learning toward the original objective, $r_t^{\mathrm{shape}}$ is an intermediate shaping reward used to mitigate reward sparsity.

Timeout penalty is additionally introduced because optimizing the original reward alone may not sufficiently enforce timely order completion during training. Shaping reward is used to provide denser feedback for intermediate decisions. We formulate shaping term as
\begin{equation}
r_t^{\mathrm{shape}}
=
\lambda_{shape} \sum_{i}
\left(
d_{i,t}^{\mathrm{tar}} - d_{i,t+1}^{\mathrm{tar}}
\right),
\end{equation}
which rewards progress toward the current delivery target and mitigates sparse delivery rewards. Importantly, $r_t^{\mathrm{timeout}}$ and $r_t^{\mathrm{shape}}$ are used only as training auxiliary signals. The final evaluation and comparison are based on Eq.~\eqref{eq:overall_profit} and the corresponding constraint-violation metrics, rather than the shaped training return.The upper-layer optimization objective is
$
R(\pi^{h}) = \mathbb{E}_{\pi^{h},\pi^{l}}
\left[
\sum_{t=0}^{T-1}\gamma^t r_t^{h}
\right].
$

\subsubsection{\textbf{Policy Network Design}}

The sensing task dispatcher is implemented as an actor--critic policy trained by PPO. Since the action is a structured UAV--sensing pair, we factorize the macro policy into two sequential categorical decisions to reduce action complexity and explicitly model UAV-task compatibility:
\begin{equation}
\pi^{h}(a_t^{h}\mid s_t^{h}) = \pi^{h}_{u}(u_i \mid s_t^{h})\, \pi^{h}_{z}(z_j \mid s_t^{h}, u_i).
\label{eq:factor_policy_final}
\end{equation}
To process the inputs, we first encode the individual UAV and sensing-task states using separate MLPs combined with the global state $g_t$. The actor network leverages a cross-attention module to extract environment-aware matching features between UAVs and tasks, alongside a permutation-invariant pooling layer to capture the global sensing summary. These fused features are fed into a UAV-selection head with masked softmax to sample the preferred UAV $u_i \sim \pi^{h}_{u}$. Conditioned on this selection, the chosen UAV's embedding is further concatenated with the fused sensing features and passed through a second actor head to sample the corresponding sensing task $z_j \sim \pi^{h}_{z}$. Finally, the critic network evaluates the overall macro state value by processing the pooled embeddings of all current UAVs and tasks through an MLP value head, providing a centralized baseline to stabilize training.

\begin{figure}[htbp]
    \centering
    \includegraphics[width=0.95\columnwidth]{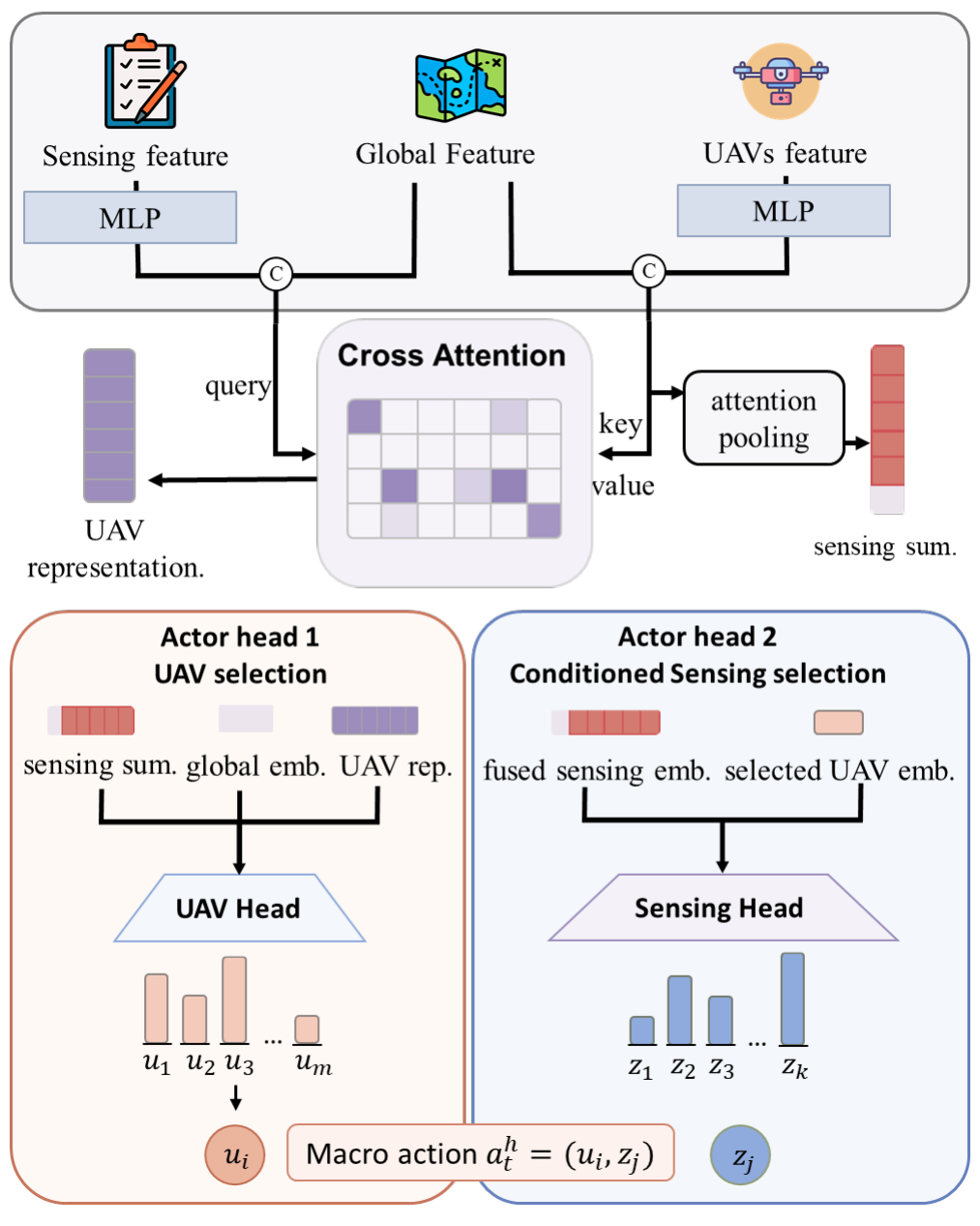}
    \caption{Sensing Dispatcher Policy Architecture}
    \label{fig:policynetwork}
\end{figure}

\subsection{Velocity Controller}

The lower layer is a velocity controller operating at the micro timescale. It is also modeled as an MDP:
\begin{equation}
\mathcal{M}^{l} = (\mathcal{S}^{l}, \mathcal{A}^{l}, \mathcal{P}^{l}, \mathcal{R}^{l}, \gamma),
\end{equation}
where $\mathcal{S}^{l}$ is the micro state space, $\mathcal{A}^{l}$ is the micro action space, $\mathcal{P}^{l}$ is the micro transition function, and $\mathcal{R}^{l}$ is the micro reward.

At each micro step $\tau$, the velocity controller observes the local UAV state and environmental context, and outputs a continuous velocity command:
\begin{equation}
a_{\tau}^{l} = v_{\tau} = \pi^{l}(s_{\tau}^{l}).
\end{equation}
The micro observation includes local motion state, target information, battery status, wind speed and wind direction. The backbone of the low-level policy is a standard MLP-based actor--critic network.

The role of the velocity controller is to execute fine-grained flight control under dynamic urban conditions. In particular, once a waypoint or task target is determined by the task manager, the lower layer is responsible for adjusting UAV speed continuously so as to balance task execution and energy consumption.

\subsection{Training Process}

As many stage-based RL frameworks did\cite{xu2020learning}, we use two-stage training instead of jointly optimizing the sensing task dispatcher and the velocity controller because joint training would make the macro-level transition dynamics non-stationary. As the lower-layer policy changes, the environment perceived by the upper layer also keeps changing. This issue is particularly severe in our two-timescale setting, where high-level combinatorial dispatch and low-level continuous control are tightly coupled. By freezing the trained velocity controller, we obtain a stable macro-level learning process and improve convergence of the sensing task dispatcher. This design achieves potential optimality for improved training stability and scalability comparing to hybrid action RL~\cite{dou2025scheduling}, which is critical in complex UAV systems.

\noindent
\paragraph{Stage I: Velocity Controller Training.}
We first train the lower-layer velocity controller $\pi^{L}$ using PPO in the micro-level environment. This stage focuses on learning stable continuous motion control under dynamic weather and battery constraints. The low-level actor and critic both use standard MLP backbones.

\noindent

\paragraph{Stage II: Sensing Task Dispatcher Training.}

At this stage, we train the high-level sensing dispatcher $\pi^h$ using PPO. The dispatcher interacts with the environment through fixed $K$-step executions of the velocity controller. 
This transforms the originally coupled macro-micro dynamics into a more stationary macro-level transition process, which significantly improves training stability for high-level sensing decisions.

Given a macro state $s_t^h$ and dispatcher action $a_t^h$, the PPO importance ratio is defined as
$\rho_t(\theta)=
\frac{\pi_{\theta}^{h}(a_t^{h}\mid s_t^{h})}
{\pi_{\theta_{\mathrm{old}}}^{h}(a_t^{h}\mid s_t^{h})}$. The clipped policy objective is 
\begin{equation}
\mathcal{L}_{\mathrm{clip}}^{h}(\theta)
=
\mathbb{E}_t
\left[
\min
\left(
\rho_t(\theta)\hat{A}_t^{h},
\;
\mathrm{clip}\big(\rho_t(\theta),1-\epsilon,1+\epsilon\big)\hat{A}_t^{h}
\right)
\right],
\end{equation}
where $\hat{A}_t^{h}$ denotes the generalized advantage estimate.The critic is optimized with the value loss $\mathcal{L}_{\mathrm{value}}^{h}(\phi)=
\mathbb{E}_t
\left[
\big(
V_{\phi}^{h}(s_t^{h})-\hat{R}_t^{h}
\big)^2
\right]$,
where $\hat{R}_t^{h}$ is the empirical return. 
To encourage exploration under dynamic sensing demands, we further introduce an entropy regularizer,
$\mathcal{L}_{\mathrm{ent}}^{h}(\theta)
=
\mathbb{E}_t
\left[
\mathcal{h}\big(\pi_{\theta}^{h}(\cdot\mid s_t^{h})\big)
\right]$.
The final optimization objective of the sensing dispatcher is
\begin{equation}
\mathcal{L}^{h}(\theta,\phi)
=
-
\mathcal{L}_{\mathrm{clip}}^{h}(\theta)
+
c_v \mathcal{L}_{\mathrm{value}}^{h}(\phi)
-
c_e \mathcal{L}_{\mathrm{ent}}^{h}(\theta),
\end{equation}
where $c_v$ and $c_e$ balance value learning and exploration regularization, respectively.

\section{Experiments}
In this section, we conduct extensive experiments to evaluate TSRL's performance in addressing SensUAV problem.
Specifically, our experiments aim to answer the following research questions (RQs):
\begin{itemize}[leftmargin=*]
    \item \textbf{RQ1}: How does TSRL perform compared to other baselines
in terms of improving system profit?
    \item \textbf{RQ2}: How do the key components of TSRL contribute to
learning effectiveness?
    \item \textbf{RQ3}: How sensitive is TSRL to key hyperparameters, specifically the potential-based reward shaping coefficient ($\lambda_{\text{shape}}$) and the two-timescale temporal abstraction scale ($K$)? 
    \item \textbf{RQ4}: What are the underlying operational behaviors and mechanisms that enable TSRL to achieve  high system profit?
\end{itemize}

\subsection{Datasets}
We preprocessed the LaDe datasets ~\cite{wu2024lade} in Hangzhou and Shanghai, and utilized logistic orders for our proposed UAV sensing problem. The information of two datasets is listed in Table 1.

\begin{itemize}[leftmargin=*]
    \item \textbf{Shanghai}: Shanghai dataset includes 3 months of last-mile delivery data from 2022/08/01 to 2022/10/31, and covers a delivery region within $[30.93^\circ \text{N}, 31.27^\circ \text{N}]$, $[121.29^\circ \text{E}, 121.63^\circ \text{E}]$ , this data includes 20707 logistic order information. We select 72 days as training dataset, 10 days as validation dataset and 10 days as test dataset. 
    \item \textbf{Hangzhou}: Hangzhou datasets includes 3 months of last-mile delivery data from 2022/05/01 to 2022/07/31, and covers a delivery region within $[30.11^\circ \text{N}, 30.44^\circ \text{N}]$, $[120.04^\circ \text{E}, 120.37^\circ \text{E}]$, this data includes 22772 logistic order information. We select 72 days as training dataset, 10 days as validation dataset and 10 days as test dataset. 
    \item  \textbf{Weather Dataset}: We collect hourly 100m wind speed and wind direction data from Open-Meteo\footnote{\url{https://doi.org/10.5281/zenodo.14582479}} with corresponding region and time mentioned above. City-level wind vector is shared by all UAVs and grid cells and updated from the hourly weather trace. 
\end{itemize}
\begin{table}[H]
  \caption{Shanghai Dataset vs. Hangzhou Dataset}
  \label{tab:freq}
  \renewcommand{\arraystretch}{1.25}
  
  \resizebox{\columnwidth}{!}{
  \begin{tabular}{llcl}
    \toprule
    \multicolumn{2}{c}{\textbf{Parameter}} & \textbf{Hangzhou} & \textbf{Shanghai} \\
    \midrule
    \multicolumn{2}{c}{Time Span} & 2022/05/01-2022/07/31 & 2022/08/01-2022/10/31 \\
    
    \multicolumn{2}{c}{Spatial Range} & 
    \begin{tabular}{@{}c@{}} $[30.11^\circ \text{N}, 30.44^\circ \text{N}]$-\\$[120.04^\circ \text{E}, 120.37^\circ \text{E}]$\end{tabular} & 
    \begin{tabular}{@{}c@{}} $[30.93^\circ \text{N}, 31.27^\circ \text{N}]$- \\ $[121.29^\circ \text{E}, 121.63^\circ \text{E}]$ \end{tabular} \\

    \multicolumn{2}{c}{Total Orders} & 22772 & 20707 \\
    
    \multicolumn{2}{c}{\begin{tabular}{@{}c@{}} Aver O-D Distance \end{tabular}} & 7.72km & 17.02km \\
    \bottomrule
  \end{tabular}
  }
\end{table}

\vspace{-2mm}
\subsection{Experimental Settings}
\begin{table*}[t]
\centering
\renewcommand{\arraystretch}{1.2} 
\setlength{\tabcolsep}{3pt} 
\begin{tabular}{llllccccccccc}
\toprule
& \multirow{2}{*}{\textbf{Method}} & \multirow{2}{*}{\textbf{Algorithms}} & \multicolumn{3}{c}{\textbf{nums=20}} & \multicolumn{3}{c}{\textbf{nums=25}} & \multicolumn{3}{c}{\textbf{nums=30}} \\
\cmidrule(lr){4-6} \cmidrule(lr){7-9} \cmidrule(lr){10-12}
& & & \textbf{Profit($\times 10^4$)} $\uparrow$ & $\Delta \uparrow$ & \textbf{T (ms)} $\downarrow$ & \textbf{Profit ($\times 10^4$)} $\uparrow$ & $\Delta \uparrow$ & \textbf{T(ms)} $\downarrow$ & \textbf{Profit($\times 10^4$)} $\uparrow$ & \textbf{$\Delta \uparrow$} & \textbf{T(ms)} $\downarrow$ \\
\midrule

\multirow{8}{*}{\rotatebox{90}{\textbf{Hangzhou}}} 
 & \multirow{3}{*}{Heuristic} 
 & Random & $2.25 \pm 0.16$ & -47.8\% & 0.04 & $2.32 \pm 0.22$ & -46.1\% & 0.04 & $2.41 \pm 0.17$ & -44.0\% & 0.04 \\
 & & DFG & $2.78 \pm 0.19$ & -35.4\% & 0.28 & $2.74 \pm 0.23$ & -36.3\% & 0.37 & $2.73 \pm 0.18$ & -36.5\% & 0.46 \\
 & & SFG & $2.83 \pm 0.19$ & -34.1\% & 0.83 & $2.80 \pm 0.25$ & -35.0\% & 0.92 & $2.79 \pm 0.18$ & -35.3\% & 1.00 \\
\cline{2-12}
 & \multirow{5}{*}{RL} 
 & SMORE & $4.28 \pm 0.04$ & -0.5\% & 4.41 & $4.04 \pm 0.42$ & -6.2\% & 4.53 & $3.83 \pm 0.48$ & -11.1\% & 4.65 \\
 & & DECO & $\underline{4.30} \pm 0.01$ & - & 3.13 & $4.14 \pm 0.28$ & -3.8\% & 3.27 & $\underline{4.31} \pm 0.01$ & - & 3.42 \\
 & & D2SN & $3.87 \pm 0.33$ & -10.0\% & 13.10 & $4.20 \pm 0.34$ & -2.5\% & 21.38 & $2.74 \pm 0.72$ & -36.3\% & 30.99 \\
 & & DyPS & $4.15 \pm 0.54$ & -3.4\% & 1.70 & $\underline{4.31} \pm 0.01$ & - & 1.76 & $3.80 \pm 0.43$ & -11.8\% & 1.83 \\
 & & \cellcolor{gray!15}\textbf{TSRL (ours)} & \cellcolor{gray!15}$\mathbf{5.07} \pm 0.07$ & \cellcolor{gray!15}\textbf{+17.9\%} & \cellcolor{gray!15}\textbf{1.29} & \cellcolor{gray!15}$\mathbf{5.04} \pm 0.04$ & \cellcolor{gray!15}\textbf{+17.0\%} & \cellcolor{gray!15}\textbf{1.31} & \cellcolor{gray!15}$\mathbf{5.40} \pm 0.40$ & \cellcolor{gray!15}\textbf{+25.2\%} & \cellcolor{gray!15}\textbf{1.33} \\

\hline\hline

\multirow{8}{*}{\rotatebox{90}{\textbf{Shanghai}}} 
 & \multirow{3}{*}{Heuristic} 
 & Random & $2.72 \pm 0.01$ & -1.9\% & 0.04 & $2.73 \pm 0.01$ & -1.8\% & 0.04 & $2.73 \pm 0.01$ & -2.0\% & 0.04 \\
 & & DFG & $2.77 \pm 0.01$ & -0.0\% & 0.20 & $2.77 \pm 0.01$ & -0.1\% & 0.26 & $2.78 \pm 0.00$ & -0.0\% & 0.33 \\
 & & SFG & $\underline{2.77} \pm 0.00$ & - & 0.94 & $\underline{2.78} \pm 0.00$ & - & 1.11 & $\underline{2.78} \pm 0.01$ & - & 1.23 \\
\cline{2-12}
 & \multirow{5}{*}{RL} 
 & SMORE & $2.47 \pm 0.43$ & -11.0\% & 4.40 & $2.76 \pm 0.06$ & -0.6\% & 4.53 & $2.49 \pm 0.45$ & -10.6\% & 4.64 \\
 & & DECO & $2.73 \pm 0.01$ & -1.5\% & 3.12 & $2.73 \pm 0.02$ & -1.7\% & 3.24 & $2.73 \pm 0.02$ & -2.0\% & 3.40 \\
 & & D2SN & $2.37 \pm 0.38$ & -14.4\% & 15.84 & $2.64 \pm 0.16$ & -5.0\% & 21.46 & $2.47 \pm 0.10$ & -11.4\% & 30.67 \\
 & & DyPS & $2.49 \pm 0.44$ & -10.4\% & 1.68 & $2.73 \pm 0.01$ & -1.8\% & 1.75 & $2.48 \pm 0.44$ & -11.0\% & 1.80 \\
 & & \cellcolor{gray!15}\textbf{TSRL (ours)} & \cellcolor{gray!15}$\mathbf{3.82} \pm 0.43$ & \cellcolor{gray!15}\textbf{+37.5\%} & \cellcolor{gray!15}\textbf{1.29} & \cellcolor{gray!15}$\mathbf{4.20} \pm 0.03$ & \cellcolor{gray!15}\textbf{+51.2\%} & \cellcolor{gray!15}\textbf{1.31} & \cellcolor{gray!15}$\mathbf{4.20} \pm 0.04$ & \cellcolor{gray!15}\textbf{+50.9\%} & \cellcolor{gray!15}\textbf{1.32} \\
\bottomrule
\end{tabular}
\caption{Profit and run time comparison under different UAV fleet sizes ($nums=20, 25, 30$) on Hangzhou and Shanghai datasets. The $\Delta$ represents the profit improvements compared to the best baseline approach (underlined). Bold and underlined digits indicate the best performance and the second-best baseline values, respectively.}
\label{tab:overall_objective_comparison}
\vspace{-15pt}
\end{table*}

\textbf{Experiment Setup}.
To evaluate our system, we build a simulation environment based on real delivery orders and weather traces. ll Each UAV's energy capacity is initialized as $1000 kJ(≈278 Wh)$, all UAVs depart from a depot and return to the depot for charging when their battery level is low; the depot location and operating region are determined from the order information in the selected delivery region. To evaluate the spatial balance of completed sensing tasks, the operating region is partitioned into a $20 \times 20$ grid. For each episode, we generate 20 sensing tasks as discrete sensing opportunities uniformly distributed within operating grid, and all compared methods share the same generated sensing task set under the same random seed. Each sensing task is associated with a time window randomly sampled from 10 to 15 hours. Each completed sensing sample is mapped to grids according to its location and completion time. The sensing profit is then computed by a hierarchical entropy function over these completed samples, which captures both the spatial balance among available sensing opportunities and the temporal balance of sensing completion events. 

\noindent
\textbf{Baselines.} Given the absence of existing literature on the hybrid-objective UAV sensing problem, we adapt several representative methods from closely related domains (e.g., order dispatching) as our baselines. These include both heuristic and reinforcement learning approaches.
Following the operational settings commonly adopted in existing UAV systems, all dispatch baselines use the city-specific maximum cruising speed. For safe terminal approach, the UAV speed is reduced to 5 km/h when approaching its current destination. This execution rule is applied uniformly across all baseline sensing-dispatch policies.
\begin{itemize}[leftmargin=*,topsep=1pt,itemsep=0pt]
    \item \textbf{Random}: When UAV is available to sensing tasks, we dispatch the sensing task randomly.
    \item \textbf{Sensing-First Greedy(SFG)}: Sensing-first greedy baseline greedily chooses the UAV–sensing pair with the minimum travel cost, allowing limited delivery interruption to prioritize sensing opportunities. 
    \item \textbf{Delivery-First Greedy(DFG)}: Delivery-first greedy baseline greedily assigns sensing tasks only to non-busy UAVs, thus preserving ongoing delivery execution. Among admissible pairs, the nearest sensing task is selected. 
    \item \textbf{SMORE~\cite{wang2024urban}}: We adapt SMORE as a two-stage transformer-pointer baseline for high-level sensing dispatch. Specifically, SMORE first encodes UAVs and sensing tasks with separate multi-head Transformer encoders, selects a UAV through a worker-level pointer module, and then selects a sensing target using a task-level pointer with a heuristic soft mask based on task utility and travel cost.
    \item \textbf{D2SN~\cite{yue2024end}}: This baseline directly scores candidate UAV--sensing pairs and selects one joint action at each decision step. In this way, it captures pairwise matching between UAV states and sensing demands within a unified encoder--decoder style architecture. This baseline serves as a learned joint-dispatch policy that contrasts with our hierarchical design.
    \item \textbf{DECO~\cite{lu2024deco}}: We adapt DECO to our UAV sensing--delivery setting as a two-stage hierarchical dispatch policy. The first stage selects a UAV, and the second stage chooses a sensing task conditioned on the selected UAV, yielding a factorized macro action. This baseline preserves the core idea of hierarchical decomposition in decision making, but applies it to UAV--sensing assignment in our system. It provides a strong hierarchical learning baseline for comparison with our method.
    \item \textbf{DyPS~\cite{wang2024dyps}}: This is a hierarchical MARL baseline that learns spatio-temporal resource allocation through dynamic parameter sharing. DyPS is used to learn the high-level sensing task dispatcher. This isolates the effect of replacing our policy with a role-modeling and group-sharing dispatcher, enabling a fair comparison on the task-allocation performance for which DyPS was originally designed.
    
\end{itemize}

\noindent
\textbf{Evaluation Metrics}.
The profit J in Equation \ref{eq:overall_profit} are used for evaluation. We also report runtime as an efficiency metric. 


\noindent 
\textbf{Training Details \& Hyperparameters}. For the Hangzhou dataset UAV max moving speed is 30km/h, for ShangHai dataset, max moving speed is set to 36km/h, the UAV numbers is explored from 20 to 30. The discount factor $\gamma$ is set to 0.99, GAE $\lambda$ is set to 0.95, clip $\epsilon$ is set to 0.2. The rollout batch size is set to 256, which is partitioned into 4 minibatches (minibatch size 64), the initial learning rate for both actor network and critic network is set to 0.0003, and the hidden size of all MLP blocks is set to 256 with token embedding dimension 128. Each micro time step is set to 5 minutes.

\noindent
\textbf{Implementations}.
We implement our algorithms in Python using PyTorch 2.3.0. All experiments were conducted on the high-performance computing platform, equipped with NVIDIA RTX A40 and RTX A800 GPUs. 
\begin{figure}[t]
    \centering
    \includegraphics[width=0.5\textwidth]{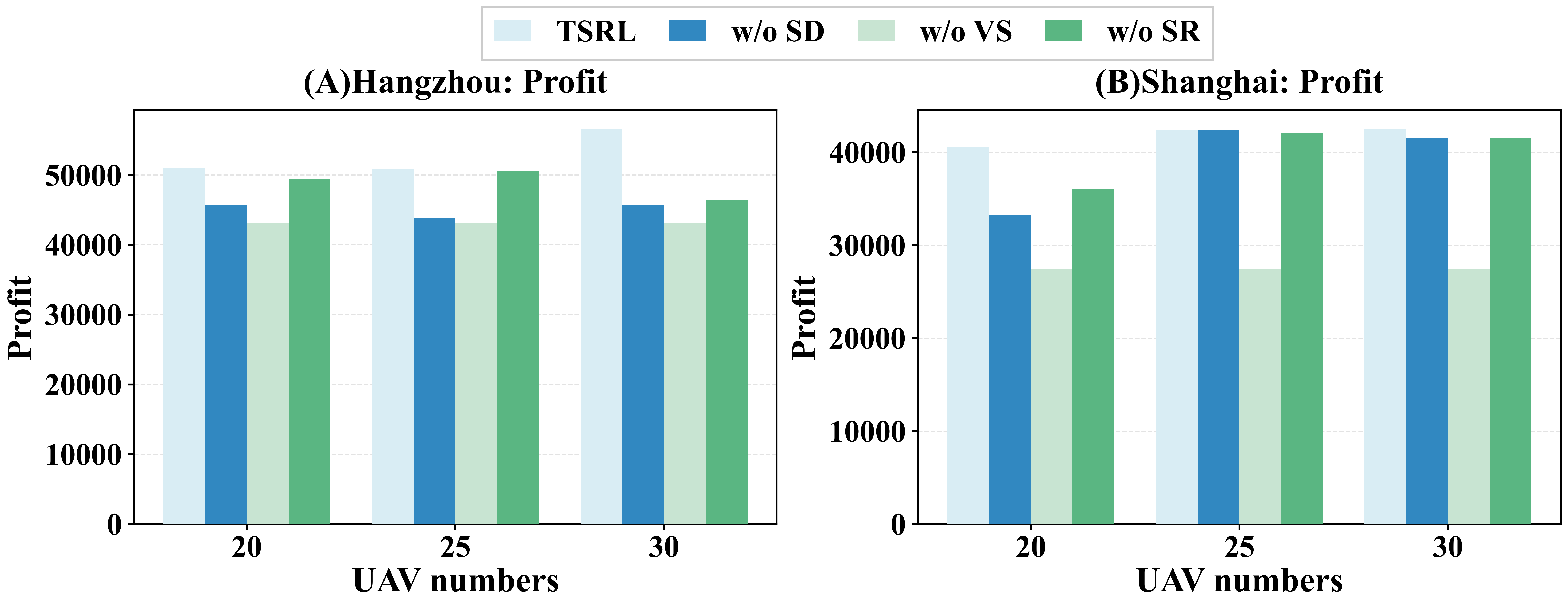}
    \caption{Ablation study on system profit.}
    \label{fig:ablation}
\end{figure}

\subsection{Overall Performance}

\textbf{Overall performance.}
Table~\ref{tab:overall_objective_comparison} compares TSRL with representative baselines under different UAV fleet sizes. TSRL consistently achieves the highest system profit across all six settings, showing its robustness under different urban scenarios and fleet scales. In Hangzhou, TSRL improves the profit over the strongest baseline by $17.9\%$, $17.0\%$, and $25.2\%$ when the number of UAVs is $20$, $25$, and $30$, respectively, with an average improvement of $20.1\%$. The advantage is even more pronounced in Shanghai, where TSRL outperforms the best competing method by $37.5\%$, $51.2\%$, and $50.9\%$, yielding an average profit improvement of $46.6\%$. Shanghai has a much longer average OD distance (17.02 km vs. 7.72 km in Hangzhou), which reduces sensing insertion flexibility and makes sensing coordination substantially more difficult. Under such long-distance delivery scenarios, several RL baselines tend to converge toward delivery-dominant strategies that rarely activate sensing tasks, while TSRL maintains effective sensing assignment through execution-aware hierarchical coordination. These results demonstrate that TSRL can better coordinate sensing dispatch and velocity control under the unified system-profit objective, rather than relying on a fixed priority rule or a weakly coupled dispatch strategy.

\noindent
\textbf{Effect of fleet size.}
Increasing the fleet size introduces additional UAV resources and higher coordination complexity. In Hangzhou, TSRL achieves the highest profit with 30 UAVs, indicating that the proposed framework can effectively utilize additional mobility resources when sufficient sensing opportunities exist. By contrast, several baselines fail to consistently benefit from larger fleets, and D2SN even exhibits severe performance degradation at 30 UAVs, suggesting that flat RL policies become increasingly unstable as the joint action space expands. In Shanghai, most baselines remain concentrated around 27K profit regardless of fleet size, while TSRL continues to improve, indicating that the key bottleneck lies in coordination quality rather than UAV quantity itself.


\noindent
\textbf{Computational Efficiency.}
TSRL also achieves strong computational efficiency during online inference, requiring only about \textbf{1.3 ms} per decision step across all settings. Compared with other RL baselines such as D2SN and SMORE, TSRL significantly reduces runtime by decomposing the original combinatorial UAV-task decision into hierarchical conditional selections instead of exhaustive joint-action evaluation. Although heuristic methods exhibit lower latency, their myopic distance-based rules ignore long-term sensing-delivery coordination and therefore produce substantially lower system profit. Overall, TSRL provides a favorable trade-off between decision quality and inference efficiency, making it practical for real-time UAV sensing systems.

\vspace{-2mm}
\subsection{Ablation Study}
To assess the contribution of each component in TSRL, we conduct an ablation study by removing one design at a time. As shown in Figure~\ref{fig:ablation}, we compare TSRL with three variants: \textbf{w/o SD}, \textbf{w/o VS}, and \textbf{w/o SR}.
\vspace{-5pt}
\begin{itemize}[leftmargin=*]
    \item \textbf{w/o SD}: We replace the proposed sensing dispatcher with a single-stage dispatcher, removing the structured UAV-task selection mechanism.
    
    \item \textbf{w/o VS}: We remove the velocity controller and use a fixed UAV flying speed as the max speed for each city during execution.

    \item \textbf{w/o SR}: We remove the shaping reward during training and only use the environment reward, including system profit and timeout penalties.
\end{itemize}
\begin{table}[htbp]
    \centering
    \caption{Ablation study on running time(ms)}
    \renewcommand{\arraystretch}{1.25}
    \label{tab:runtime}
    \begin{tabular}{llccc}
        \toprule
        \textbf{Dataset} & \textbf{Method} & \textbf{nums=20} & \textbf{nums=25} & \textbf{nums=30} \\
        \midrule
        \multirow{2}{*}{Hangzhou} 
        & TSRL (ours) & \textbf{1.29} & \textbf{1.31} & \textbf{1.33} \\
        & w/o SD & 1.55 & 1.77 & 1.98 \\
        \midrule
        \multirow{2}{*}{Shanghai} 
        & TSRL (ours) & \textbf{1.29} & \textbf{1.31} & \textbf{1.32} \\
        & w/o SD & 1.54 & 1.75 & 1.97 \\
        \bottomrule
    \end{tabular}
    
\end{table}

As shown in Figure~\ref{fig:ablation}, removing any component generally degrades TSRL, confirming that the performance gain does not come from a single isolated module but from the joint design of structured dispatch, velocity scheduling, and reward shaping. Among these variants, removing the velocity controller leads to the most severe profit drop in most cases, especially in Shanghai, where the profit collapses close to the level of rule-based baselines. This verifies that adaptive velocity control is crucial for reducing execution cost and improving the timing of task completion under dynamic urban conditions. Removing the sensing dispatcher also weakens performance because the policy can no longer explicitly model UAV--task compatibility. Removing reward shaping causes a smaller but still visible degradation, indicating that dense guidance helps stabilize training under sparse profit feedback. Overall, the ablation results show that TSRL's advantage comes from coupling high-level dispatch with low-level execution rather than optimizing them independently. Beyond profit improvements, Table~\ref{tab:runtime} further reveals that our structured sensing dispatcher accelerates online inference compared to the single-stage variant (\textbf{w/o SD}). This confirms that factorizing the combinatorial action space not only captures UAV-task compatibility more accurately but also significantly reduces computational overhead.

\subsection{Hyperparameter Study}
\begin{figure}
    \centering
    \includegraphics[width=0.5\textwidth]{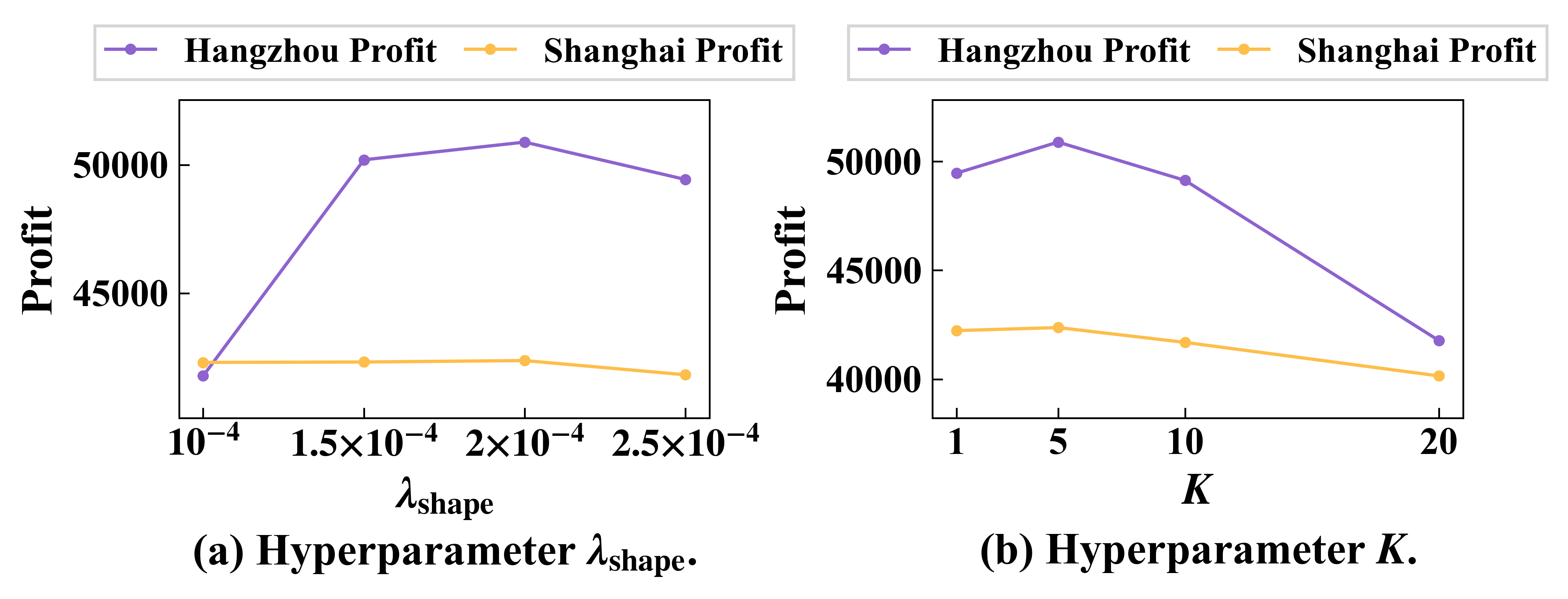}
    \caption{Hyperparameter study on system profit.}
    \label{fig:hyperparameter_profit}
\end{figure}

\noindent
\textbf{Impact of $\lambda_{\text{shape}}$.}
Figure~\ref{fig:hyperparameter_profit}(a) reports the sensitivity of TSRL to the shaping coefficient $\lambda_{\text{shape}}$, which controls the strength of the potential-based shaping term. This parameter determines the relative strength between the sparse system-profit signal and the dense geometric guidance toward unfinished sensing targets. When $\lambda_{\text{shape}}$ is too small, the shaping signal is insufficient to guide high-level dispatch before task completion, resulting in weaker policy learning. This explains the clear performance drop in Hangzhou at $\lambda_{\text{shape}}=10^{-4}$. As $\lambda_{\text{shape}}$ increases to $2\times10^{-4}$, TSRL reaches its best profit, indicating that the shaping signal is well aligned with the system-profit objective. However, further increasing $\lambda_{\text{shape}}$ does not bring additional gains and slightly reduces profit, suggesting that an overly strong shaping term may bias the dispatcher toward local proximity rather than global profit maximization. The milder variation in Shanghai indicates that its task dynamics are less sensitive to shaping strength. Therefore, we set $\lambda_{\text{shape}}=2\times10^{-4}$ in the main experiments.

\noindent
\textbf{Impact of $K$.}
Figure~\ref{fig:hyperparameter_profit}(b) studies the effect of $K$, the number of low-level velocity-control steps executed between two consecutive high-level dispatch decisions. In TSRL, $K$ controls the effective holding time of a high-level UAV--task assignment and directly reflects the two-timescale design. A small $K$ makes the dispatcher highly reactive, but it may revise assignments before the selected UAV has made meaningful progress toward the target. In contrast, an excessively large $K$ makes dispatch decisions stale, reducing the system's ability to respond to newly arriving orders, changing wind conditions, and expiring sensing opportunities. The results show that $K=5$ achieves the best profit on both datasets, confirming that TSRL benefits from an intermediate temporal abstraction. The sharper degradation in Hangzhou when $K$ becomes large suggests that dense urban demands impose a stronger penalty on delayed dispatch updates, whereas Shanghai is less sensitive due to relatively smoother task dynamics. We therefore set $K=5$ for all main experiments.

\begin{figure}[t]
    \centering
    \includegraphics[width=0.5\textwidth]{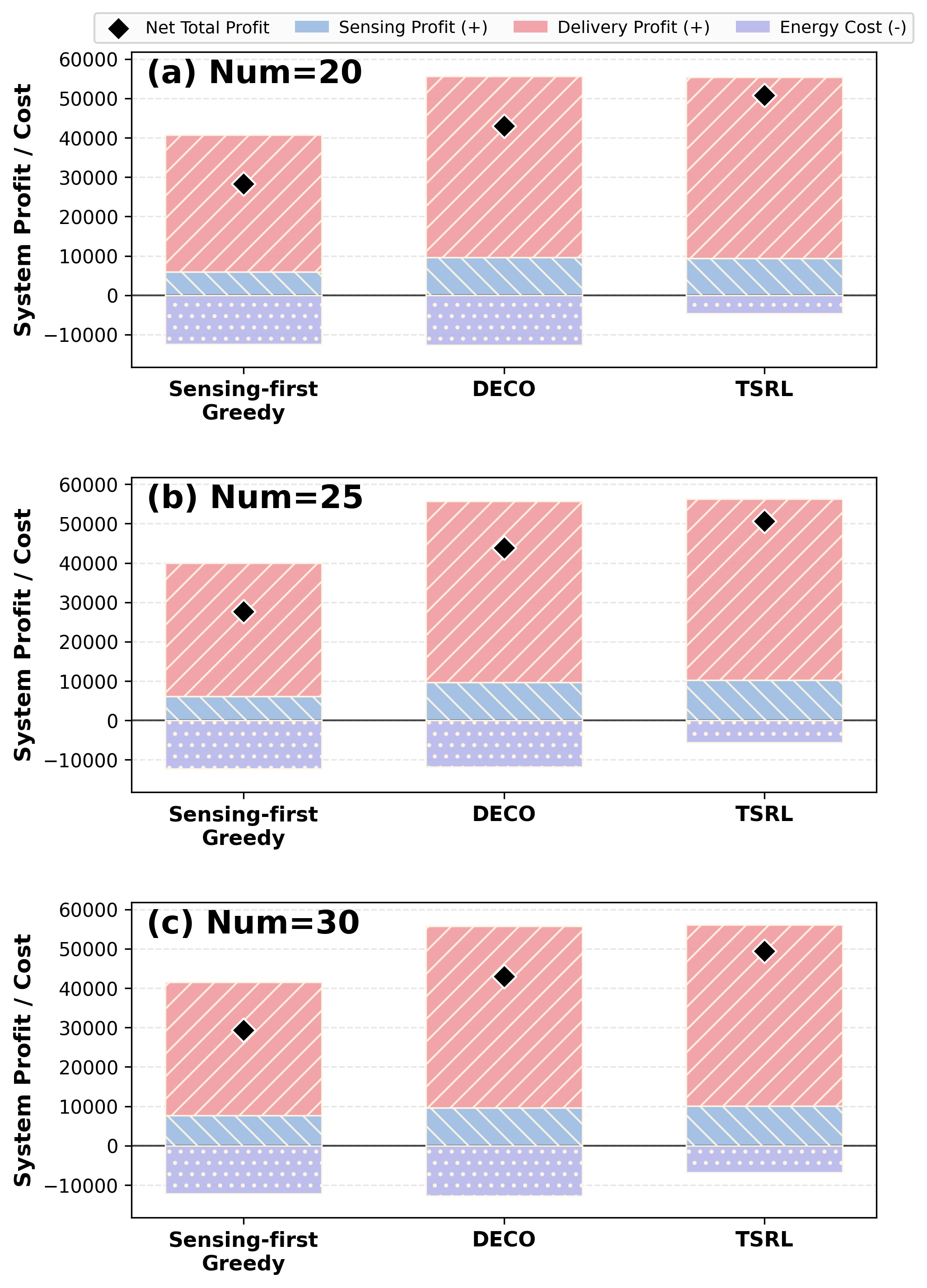}
    \caption{System behavior analysis.}
    \label{fig:system_behavior}
\end{figure}

\subsection{System Behavior Analysis}

To reveal the operational mechanisms of our model and verify how the maximum profit is attained, we evaluate TSRL against the best heuristic (Sensing-first Greedy) and reinforcement learning baselines (DECO). The comparative results are presented in Figure~\ref{fig:system_behavior}.

As shown in Figure~\ref{fig:system_behavior}, Sensing-first Greedy obtains relatively low net system profit because it overly prioritizes sensing task and consequently sacrifices delivery completion, leading to lower delivery profit. In contrast, TSRL and DECO achieve higher delivery profit through more reasonable sensing dispatch, indicating that effective sensing coordination should not come at the expense of delivery service. The remaining profit gap between TSRL and DECO mainly comes from concurrent sensing allocation and lower-level energy efficiency. Specifically, TSRL demonstrates a substantial advantage over the greedy baseline in sensing task assignment, validating the superior scheduling capability of our sensing dispatcher. Meanwhile, by adaptively optimizing lower-level execution speeds, our velocity controller substantially reduces energy consumption compared to DECO. In summary, these behavioral insights confirm that instead of improving sensing profit by sacrificing delivery tasks, the synergistic design of TSRL achieves high task throughput and efficient energy saving.

\vspace{-4mm}
\subsection{Case Study}
To intuitively visualize how TSRL differs from the baselines in making dispatch decisions, we plot the trajectory of the same UAV under different methods, together with the completed and unfinished sensing tasks. 

\begin{figure}[H]
    \centering    \includegraphics[width=0.5\textwidth]{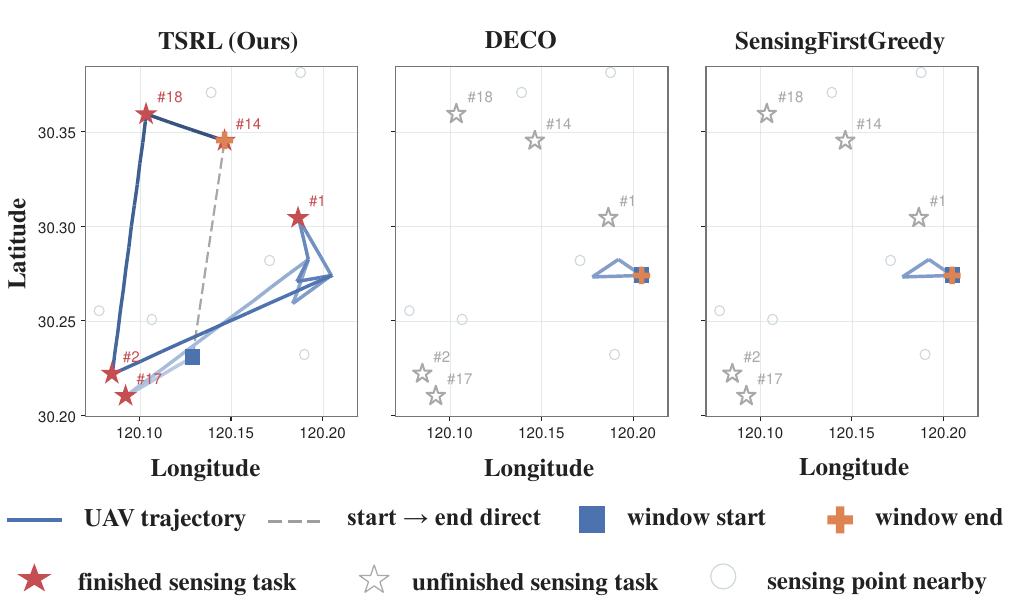}
    \caption{Case study.}
    \label{fig:case study}
\end{figure}

 Figure~\ref{fig:case study} visualizes the fleet-level scheduling difference between TSRL and the baselines for UAV 20 on 2022/05/23. TSRL identifies UAV 20 as free from delivery duties and dispatches it on a multi-stop sensing tour, completing five sensing tasks. In contrast, DECO and SFG leave the same UAV stationary for most of the window, implicitly relying on UAVs already serving delivery orders to collect sensing data opportunistically. This strategy may systematically fail to cover peripheral sensing points that lie far from major delivery corridors.

\vspace{-3mm}
\section{Related Works}
\subsection{Participatory Sensing for Delivery Drones}

UAVs equipped with cameras, LiDARs, or other sensors act as flying base stations and data collectors, which greatly enhance situational awareness in cities ~\cite{rashid2020socialdrone}. They can patrol areas to monitor traffic, detect incidents, and collect environmental data ~\cite{rashid2020socialdrone, dou2025scheduling}. UAV crowdsensing also enables flexible coverage for events or disasters, complementing ground sensor networks. The UAV deployment problem is often split into static  deploying hovering UAVs to cover fixed areas versus dynamic trajectory planning which routing UAVs over time to follow moving phenomena ~\cite{lai2024hyperspectral}. Static placement fails to adapt when targets or demand areas move, so modern research emphasizes trajectory design. A key insight is that mobile UAVs can reduce their distance to sensing targets, improving data quality or reducing communication delays ~\cite{mahmood2025uav}.

Optimization of UAV trajectories is inherently multi-objective. Typical goals include maximizing sensing coverage or utility while minimizing costs such as total mission time, energy consumption, or transmission power ~\cite{ji2023survey, dou2025scheduling}. For example, some formulations maximize the total amount of high-quality imagery collected subject to UAV battery limits, while others minimize estimation error by servicing more ground sensors. Many works extend this formulation to joint delivery-and-sensing scenarios. Xiang et al. ~\cite{xiang2021reusing} reuse delivery drones during their flights, jointly optimizing route, and sensing duration to maximize combined utility, resulting in non-convex programs solved via approximation algorithms. 


However, most existing studies treat sensing as standalone assignment or route planning, overlooking its integration with dynamic logistics. They also rely on simplified energy models with fixed speeds, neglecting wind-induced energy variations and treating velocity as an execution detail rather than a decision variable.

\subsection{\mbox{Reinforcement Learning for UAV Scheduling}}

To overcome the limitations of traditional optimization, some \mbox{studies} formulate UAV scheduling as a sequential decision-making \mbox{problem} and apply Reinforcement Learning to learn adaptive policies ~\cite{dou2025scheduling}. In an RL framework, UAVs interact with the environment modeled as a Markov decision process , where states encode positions, energy levels, and task demands, and actions determine routing and service decisions. Deep RL methods such as DQN ~\cite{van2016deep}, DDPG ~\cite{lillicrap2015continuous}, and PPO ~\cite{schulman2017proximal} have been widely used to handle continuous and high-dimensional control problems. For example, Qin et al. ~\cite{qin2025physics} incorporate weather effects into an MDP formulation and optimize UAV trajectories under rain using Double Q-learning. 

Multi-agent RL further extends this paradigm to coordinated multi-UAV systems. A common approach is centralized training with decentralized execution ~\cite{amato2024introduction}. Xu et al. ~\cite{xu2024scalable} group UAV agents to share policies and improve scalability, while Dou et al. ~\cite{dou2025scheduling} propose a hybrid-action RL framework that jointly models discrete decisions  and continuous controls. DeliverSense ~\cite{chen2022deliversense} and PathPool ~\cite{xiang2025path} demonstrate that RL can outperform heuristic baselines.

Despite these advances, most RL-based methods use flat policies that jointly optimize sensing and routing, leading to large action spaces and unstable training. They also overlook UAVs’ multi-timescale dynamics and often treat velocity implicitly, motivating multi-stage frameworks with decoupled decisions and wind-aware energy modeling.

\section{Conclusion \& Future Work}

In this paper, we study participatory sensing for delivery drones in dynamic urban environments, where UAVs jointly serve delivery orders and sensing tasks under wind-aware energy consumption. We formulate the \textbf{SensUAV} problem with discrete UAV-task assignment and continuous velocity scheduling, and prove its NP-hardness. We propose \textbf{TSRL}, a two-stage two-timescale reinforcement learning framework that combines a low-level velocity controller with a factorized high-level sensing dispatcher. Experiments on two real-world order datasets show that TSRL consistently outperforms heuristic and learning-based baselines in system profit and efficiency. Future work will incorporate finer-grained aerodynamic modeling, and real-world deployment constraints.

\section{GenAI Usage Disclosure}

We used a large language model (Gemini 3.1 Pro) to assist with grammar editing of this manuscript. The LLM was not used to generate scientific content, results, or conclusions. All technical content and claims were produced and verified by the authors.

\begin{acks}
This work is supported by the National Natural Science Foundation of China (No. 62306033, No. 62402414), the Guangdong Basic and Applied Basic Research Foundation (No. 2025A1515011994), Guangdong Provincial Project 2025D03J0014, Guangzhou Municipal Science and Technology Project (No. 2023A03J0011), the Guangzhou Industrial Information and Intelligent Key Laboratory Project (No. 2024A03J0628), Guangdong Provincial Key Lab of Integrated Communication, Sensing and Computation for Ubiquitous Internet of Things (No. 2023B1212010007) and the Research Travel Grant of the Base of Red Bird MPhil (RBM) at The Hong Kong University of Science and Technology (Guangzhou). We would like to thank our RBM Project Supervisor Qianqian Wei for the academic support.
\end{acks}

\clearpage
\bibliographystyle{ACM-Reference-Format}
\bibliography{sample-base}


\appendix
\section{Training Process}

\begin{algorithm}[H]
\caption{Two-Stage Training Process of TSRL}
\label{alg:two_level_training}
\begin{algorithmic}[1]
\State \textbf{Input}: Environment $env$; macro rollout length $B^h$; micro rollout length $B^l$; total steps $T^h_{total},T^l_{total}$; micro horizon $K$.
\State \textbf{Output}: Trained dispatcher $\pi_\theta^h$ and velocity controller $\pi_\phi^l$.

\Statex \Comment{\textbf{Stage I: Train velocity controller}}
\State Initialize $\pi_\phi^l$;
\For{$u=1$ \textbf{to} $\lfloor T^l_{total}/B^l \rfloor$}
    \State Collect $B^l$ micro transitions with $\pi_\phi^l$;
    \State Update $\phi$ using PPO with GAE;
\EndFor

\Statex \Comment{\textbf{Stage II: Train sensing dispatcher}}
\State Initialize $\pi_\theta^h$ and set $step_h \leftarrow 0$;
\While{$step_h < T^h_{total}$}
    \State Empty macro buffer $\mathcal{B}^h$;
    \While{$|\mathcal{B}^h| < B^h$}
        \State Observe macro state $s_t^h$ and sample $a_t^h \sim \pi_\theta^h(\cdot|s_t^h)$;
        \State Apply $a_t^h$ to update the sensing assignment;
        \State $R_t^h \leftarrow 0$;        
        \For{$k=1$ \textbf{to} $K$}
            \State Execute $a_{t,k}^l \sim \pi_\phi^l(\cdot|s_{t,k}^l)$;
            \State Step $env$ to update state $s_t^h$;
            \If{episode terminates}
                \State \textbf{break};
            \EndIf
        \EndFor
        \State Observe $s_{t+1}^h$ and compute $R_t^h$ by Eq.~\eqref{eq:macro_reward_final};
                \State Store $(s_t^h,a_t^h,R_t^h,s_{t+1}^h,done_t^h)$ in $\mathcal{B}^h$;
        \State $step_h \leftarrow step_h+1$;
        \If{$done_t^h$} \State Reset $env$; \EndIf
    \EndWhile
    \State Update $\theta$ using PPO with GAE on $\mathcal{B}^h$;
\EndWhile
\State \textbf{return} $\pi_\theta^h,\pi_\phi^l$;
\end{algorithmic}
\end{algorithm}

\section{Real-World Deployment Visualization}
\begin{figure}[H]  
    \centering      
    \includegraphics[width=0.5\textwidth]{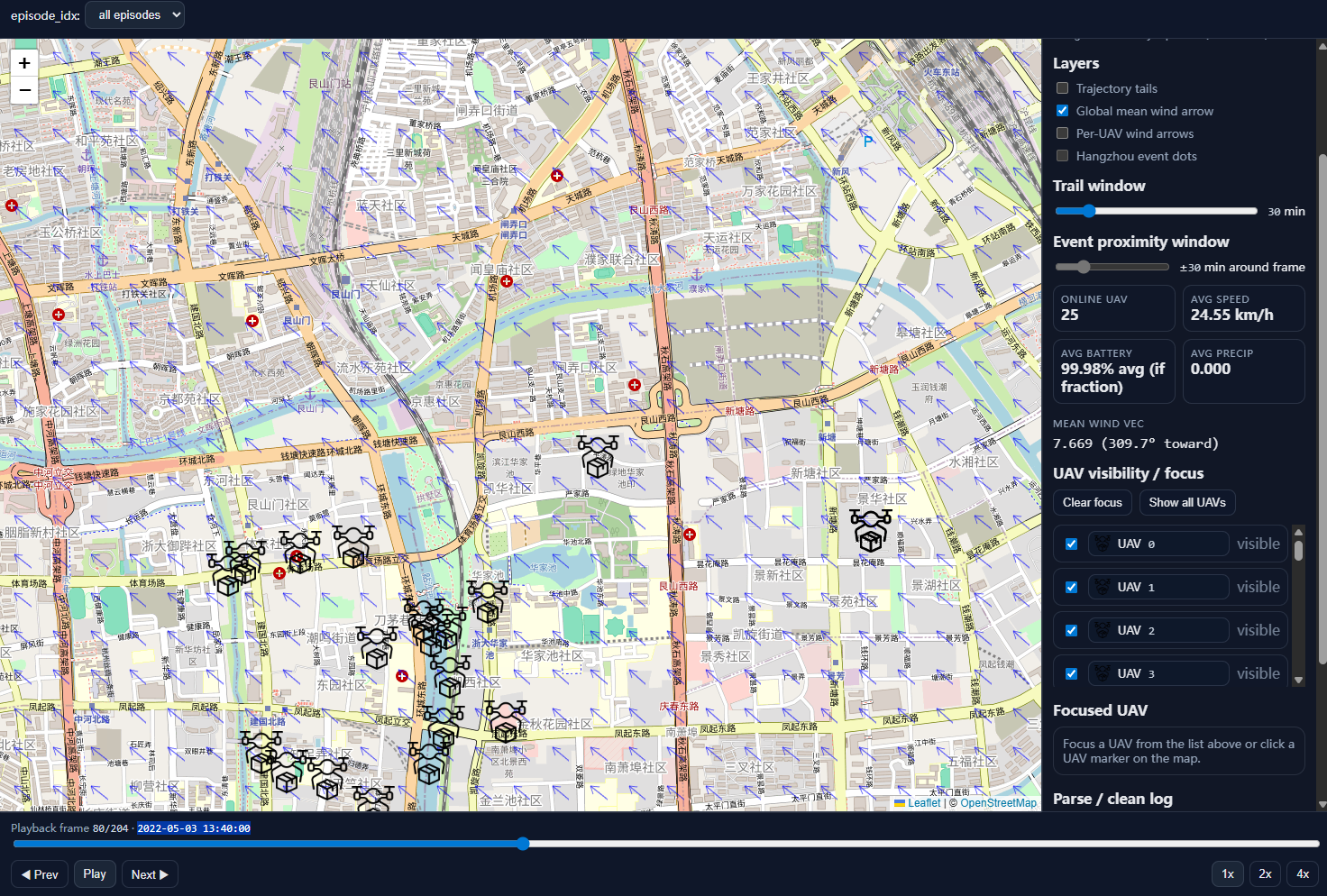}  
    \caption{Web Platform of SensUAV system}  
    \label{fig:5}     
\end{figure}

To better understand how TSRL behaves after deployment, we build a lightweight visualization platform using the LaDe test set. The platform places the replayed UAV trajectories on an OpenStreetMap background and shows how the agents move as sensing and delivery demands change over time. We use ten test days from LaDe, so that the replay covers different demand patterns and weather conditions. In each replay, the trained TSRL policy determines the UAV movements and task choices under the SensUAV setting, where each UAV must coordinate delivery and sensing while respecting operational constraints such as battery usage and wind. The platform records the position of each UAV at every time step and visualizes its recent trajectory, remaining battery, and local wind condition. This gives a direct qualitative view of how TSRL handles the coupled sensing-and-delivery process, and complements the quantitative results by showing that the learned policy can produce coherent multi-UAV coordination on real geographic data. Our source codes are available in the supplementary materials.

\end{document}